\documentclass[preprint]{article} 

\usepackage{aaai2027}  
\usepackage[hyphens]{url}  
\usepackage{graphicx} 
\usepackage{natbib}  
\usepackage{caption} 
\usepackage{algorithm}
\usepackage{algorithmic}

\usepackage{newfloat}
\usepackage{listings}
\DeclareCaptionStyle{ruled}{labelfont=normalfont,labelsep=colon,strut=off} 
\floatstyle{ruled}
\newfloat{listing}{tb}{lst}{}
\floatname{listing}{Listing}

\usepackage{booktabs}

\usepackage{multirow}
\usepackage{makecell}
\usepackage{amssymb}
\usepackage{adjustbox}
\usepackage{xcolor}
\usepackage{array}
\usepackage{pifont}
\newcommand{\cmark}{\checkmark}
\newcommand{\xmark}{\ding{55}}
\newcommand{\modellogo}[2]{%
  \makebox[1.6em][c]{%
    \raisebox{-0.15em}{%
      \includegraphics[height=1.05em,keepaspectratio]{#1}%
    }%
  }\,#2%
}

\title{Can Multimodal Large Language Models Understand OCT?}

\author{
    Baochen Fu\textsuperscript{\rm 1,\rm 2},
    Wenzhi Deng\textsuperscript{\rm 1},
    Baihao Jin\textsuperscript{\rm 1},
    Yang Li\textsuperscript{\rm 4},\\
    Zihan Nie\textsuperscript{\rm 1},
    Kailin Jiang\textsuperscript{\rm 5},
    Yuntao Du\textsuperscript{\rm 1,\rm 2,\rm 3}\corresponding,
    Weiye Song\textsuperscript{\rm 1}\corresponding
}
\affiliations {
    \textsuperscript{\rm 1}Shandong University\\
    \textsuperscript{\rm 2}Joint SDU-NTU Centre for Artificial Intelligence Research\\
    \textsuperscript{\rm 3}State Key Laboratory for Novel Software Technology\\
    \textsuperscript{\rm 4}Shandong Provincial Qianfoshan Hospital\\
    \textsuperscript{\rm 5}University of Science and Technology of China
}

\begin{document}

\maketitle

\begin{abstract}
Optical coherence tomography (OCT) imaging is essential for the diagnosis and treatment of retinal diseases. Although multimodal large language models (MLLMs) have demonstrated considerable potential in medical image analysis, existing benchmarks largely reduce OCT understanding to coarse-grained disease classification or isolated visual question answering, leaving the complete cognitive process from visual perception to clinical reasoning insufficiently evaluated. To address this limitation, we introduce \textbf{OCT-Bench}, a comprehensive benchmark dedicated to OCT image understanding. OCT-Bench comprises 10,076 high-quality multiple-choice questions constructed from 4,137 OCT images across seven public datasets. Following the real-world clinical interpretation workflow, we establish a hierarchical capability taxonomy consisting of 20 fine-grained tasks across three dimensions: Perception, Cognition, and Reasoning. These tasks cover a broad range of capabilities, including imaging attributes, retinal anatomy, lesion characteristics, spatial relationships, disease assessment, therapeutic decision-making, and prognostic management. We systematically evaluate 20 representative MLLMs, including proprietary models, open-source general-purpose models, and medical-domain models. Experimental results demonstrate that current models remain substantially short of reliable OCT understanding. Moreover, neither medical-domain adaptation nor increased model scale consistently improves performance across capability levels. OCT-Bench enables comprehensive and fine-grained evaluation of MLLMs, providing a foundation for identifying capability bottlenecks and advancing clinically grounded OCT understanding.
\end{abstract}
\begin{links}
    \link{Code}{https://github.com/baochenfu/OCT-Bench/}
\end{links}

\section{Introduction}
Multimodal large language models (MLLMs) have recently achieved remarkable progress on general-purpose vision tasks~\cite{wu2024visionllm,caffagni2024revolution,qi2025context}, driven by increasingly powerful visual understanding and cross-modal reasoning capabilities. This progress has also revealed considerable potential for assisting medical image analysis. Unlike natural-image understanding, however, medical image interpretation requires more than recognizing visual patterns~\cite{bai2025label,lin2025taming}: models must integrate specialized medical knowledge to analyze disease, form clinical judgments, and support decision-making. This raises a fundamental question: \emph{can current MLLMs genuinely understand complex medical images and complete the full process from visual perception to clinical reasoning?}

Optical coherence tomography (OCT), one of the most important imaging techniques in ophthalmic practice~\cite{huang1991optical}, provides high-resolution cross-sectional views of retinal structures and is widely used for disease diagnosis, treatment planning, and longitudinal follow-up~\cite{gurumoorthy2025role,fang2025research}. OCT interpretation nevertheless presents a substantial technical and clinical challenge because scans contain complex layered anatomy, subtle variations in tissue reflectivity, and frequently coexisting abnormalities. In routine practice, clinicians first assess image quality and identify retinal anatomy, then examine lesion morphology, spatial relationships, and pathological characteristics, and finally integrate these findings with clinical knowledge to establish a diagnosis, determine treatment, and evaluate prognosis~\cite{wang2024advances,chen2024visual}. This progression from low-level visual perception to high-level clinical reasoning makes OCT a particularly demanding test bed for evaluating medical image understanding in MLLMs.

Existing benchmarks, however, remain insufficient for comprehensively evaluating OCT understanding in two important respects. First, general-purpose multimodal benchmarks predominantly focus on natural images and generic visual capabilities~\cite{jiang2026large,fu2026mmku,liu2026amo}, offering little systematic evaluation of medical images and OCT in particular. Existing ophthalmic benchmarks~\cite{zou2025benchmarking,srinivasan2026benchmarking} primarily address textual medical knowledge, fundus imagery, retinal image enhancement, or ophthalmic surgical videos. Although LMOD includes OCT among several ophthalmic modalities~\cite{qin2025lmod}, its evaluation remains limited to a small number of coarse-grained tasks and cannot capture the diverse capabilities required for clinical OCT interpretation.

\begin{figure*}[ht]
\centering
\includegraphics[width=\textwidth]{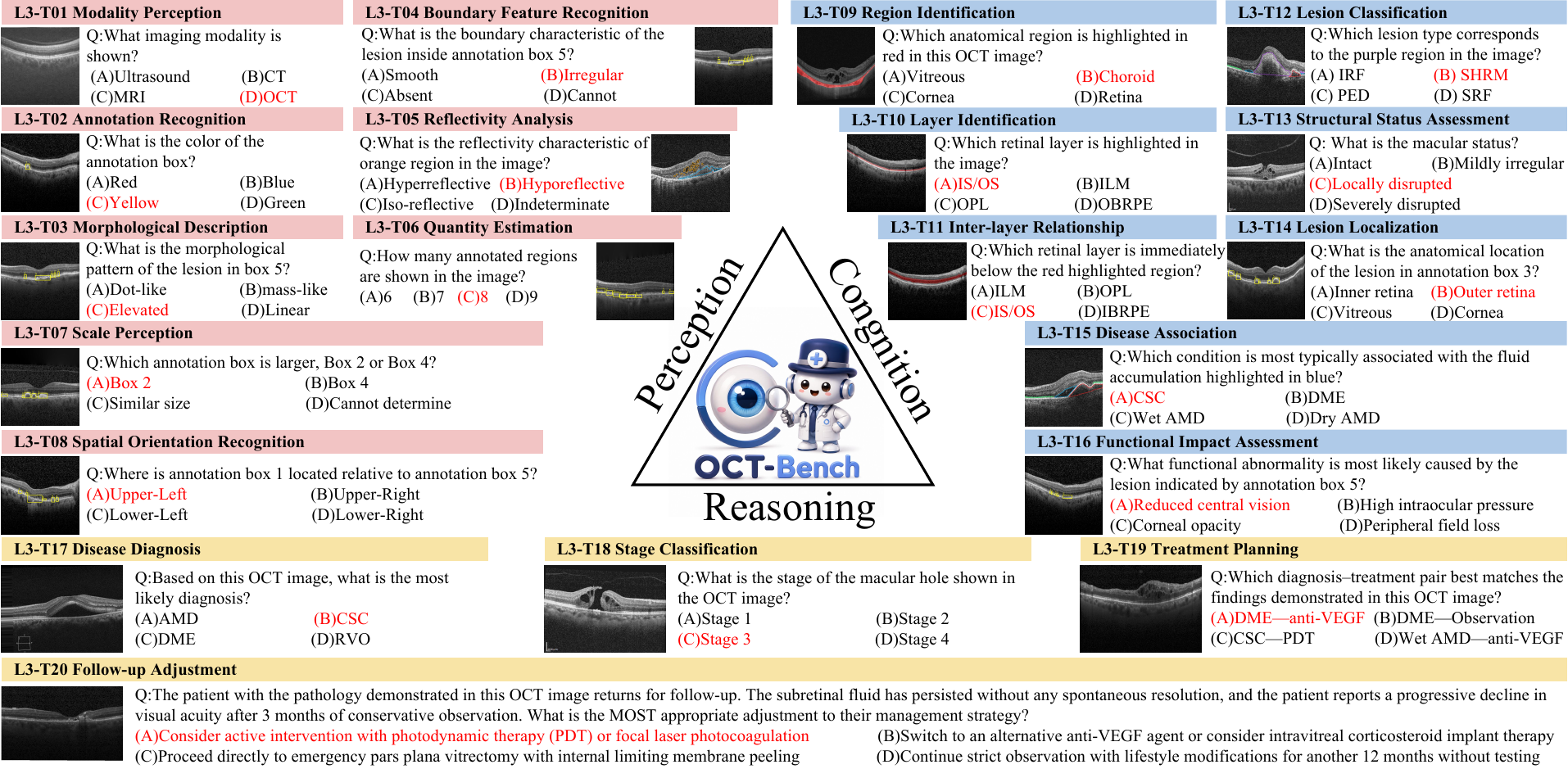}
\caption{Overview of OCT-Bench.}
\label{fig:fig3}
\end{figure*}

Second, and more importantly, existing benchmarks commonly reduce OCT understanding to disease classification or isolated visual question answering, overlooking the hierarchical cognitive process underlying real-world interpretation~\cite{chen2026visual}. OCT analysis does not proceed directly from an image to a diagnosis; rather, it follows a progressive pathway from visual perception through medical cognition to clinical reasoning. Existing evaluations often conflate these levels within a single task. When a model makes an incorrect prediction, it is therefore difficult to determine whether the failure arises from inadequate visual perception, deficient medical understanding, or unreliable clinical reasoning\cite{zhou2026drvd}. This entanglement not only hinders fair and informative model comparison but also provides limited guidance for targeted model improvement.

To address these limitations, we introduce OCT-Bench (Figure~\ref{fig:fig3}), a comprehensive benchmark for evaluating MLLMs on OCT image understanding. Rather than treating OCT interpretation as a conventional disease-classification problem, OCT-Bench follows the clinical interpretation workflow and establishes a hierarchical taxonomy comprising three primary dimensions, Perception, Cognition, and Reasoning, together with nine capability groups and 20 fine-grained tasks, as illustrated in Figure~\ref{fig:fig1}. Built from 4,137 carefully curated and quality-controlled OCT images collected from seven public datasets, OCT-Bench contains 10,076 expert-verified high-quality multiple-choice questions covering imaging quality, retinal anatomy, lesion characteristics, spatial relationships, disease assessment, therapeutic decision-making, and prognostic management. Compared with existing benchmarks (Table~\ref{table1}), OCT-Bench supports both comprehensive assessment of overall performance and precise diagnosis of capability bottlenecks across different stages, revealing how errors may propagate from low-level visual perception to high-level clinical reasoning.

Based on OCT-Bench, we systematically evaluate 20 representative MLLMs, including proprietary models, open-source general-purpose models, and medical-domain models. Our results demonstrate that current MLLMs remain far from reliable OCT understanding. The best-performing model achieves only 62.0\% overall accuracy, and performance declines markedly as the required capability advances: the highest perception score reaches 75.8\%, whereas the highest reasoning score is only 42.9\%. Moreover, neither medical-domain adaptation nor increased model scale yields consistent improvements across capability levels. These results indicate that current MLLMs still struggle to perform reliable clinical reasoning grounded in OCT images.

The main contributions are summarized as follows:
\begin{itemize}
    \item We introduce OCT-Bench, a large-scale benchmark dedicated to OCT image understanding, containing 10,076 questions over 4,137 images from seven public datasets and enabling comprehensive assessment beyond coarse disease classification.
    \item We develop a clinically grounded hierarchical taxonomy that decomposes OCT understanding into three primary dimensions, nine capability groups, and 20 fine-grained tasks, allowing capability deficiencies at different stages to be precisely identified and analyzed.
    \item We systematically evaluate 20 representative MLLMs and reveal substantial performance degradation from visual perception to clinical reasoning, while providing a fine-grained analysis of the strengths and limitations of different model families.
\end{itemize}

\begin{table*}[ht]
\centering
\resizebox{\textwidth}{!}{%
\begin{tabular}{lccccccccc}
\toprule
\textbf{Benchmarks} & \textbf{OCT} & \textbf{\makecell{Multimodal}} & \textbf{\makecell{Disease\\Classification}} & \textbf{\makecell{Anatomy\\Recognition}} & \textbf{\makecell{Lesion\\Recognition}} & \textbf{\makecell{Visual\\Perception}} & \textbf{\makecell{Medical\\Cognition}} & \textbf{\makecell{Clinical\\Reasoning}} & \textbf{Tasks} \\
\midrule
\multicolumn{10}{c}{\textbf{General-Domain Benchmarks}} \\
\midrule
MMBench                 & \xmark & \cmark & \xmark & \xmark & \xmark & \cmark & \xmark & \xmark & 20 \\
MME-RealWorld             & \xmark & \cmark & \xmark & \xmark & \xmark & \cmark & \xmark & \xmark & 43 \\
UNK-VQA                   & \xmark & \cmark & \xmark & \xmark & \xmark & \cmark & \xmark & \xmark & 5 \\
MMCBench                  & \xmark & \cmark & \xmark & \xmark & \xmark & \cmark & \xmark & \xmark & 4 \\
MathVista                 & \xmark & \cmark & \xmark & \xmark & \xmark & \cmark & \xmark & \xmark & 6 \\
SEED-Bench                & \xmark & \cmark & \xmark & \xmark & \xmark & \cmark & \xmark & \xmark & 12 \\

\midrule
\multicolumn{10}{c}{\textbf{Ophthalmology-Specific Benchmarks}} \\
\midrule
Eval-GPT-Ophth            & \xmark & \xmark & \xmark & \xmark & \xmark & \xmark & \cmark & \xmark & 13 \\
Bench-Myopia              & \xmark & \xmark & \xmark & \xmark & \xmark & \xmark & \cmark & \cmark & 6 \\
EyeBench                  & \xmark & \xmark & \xmark & \xmark & \cmark\rlap{2} & \cmark & \xmark & \xmark & 7 \\
OphNet                    & \xmark & \cmark & \xmark & \xmark & \xmark & \xmark & \cmark & \xmark & 4 \\
LMOD                      & \cmark & \cmark & \cmark\rlap{2} & \cmark\rlap{2} & \cmark\rlap{2} & \xmark & \xmark & \xmark & 3 \\
\textbf{OCT-Bench (Ours)} & \textbf{\cmark} & \textbf{\cmark} & \textbf{\cmark\rlap{10}} & \textbf{\cmark\rlap{7}} & \textbf{\cmark\rlap{10}} & \textbf{\cmark} & \textbf{\cmark} & \textbf{\cmark} & \textbf{20} \\
\bottomrule
\end{tabular}%
}
\caption{Comparison of OCT-Bench with existing general-domain and ophthalmology-specific benchmarks.}
\label{table1}
\end{table*}

\section{Related Work}

\subsection{Multimodal Large Language Models}

Multimodal large language models (MLLMs) integrate visual encoders with large language models to enable image-grounded understanding and reasoning. Recent proprietary models, such as GPT-4o, Gemini, and Grok, have demonstrated strong multimodal capabilities, while open-source models, including LLaVA, Phi-3-Vision, InternVL, Qwen2.5-VL, and Gemma 3, have substantially advanced multimodal research through improved reproducibility and transparency~\cite{liu2024improved,li2024llava,bilenko2024new,chen2024expanding,bai2025qwen2,team2024gemma}. Medical MLLMs, such as HealthGPT, MedGemma, Lingshu, Hulu-Med, MediX-R1, and Fleming-VL, further leverage biomedical knowledge and medical image-text data to enhance medical image understanding and clinical reasoning. However, most existing MLLMs are trained on natural images or broad medical corpora, whereas OCT interpretation requires understanding fine-grained retinal structures and subtle pathological features. Whether current proprietary, open-source, and medical MLLMs can effectively understand and reason over OCT images therefore remains an open question.

\subsection{Ophthalmic Evaluation Benchmarks}

\begin{figure}[ht]
\centering
\includegraphics[width=0.95\columnwidth]{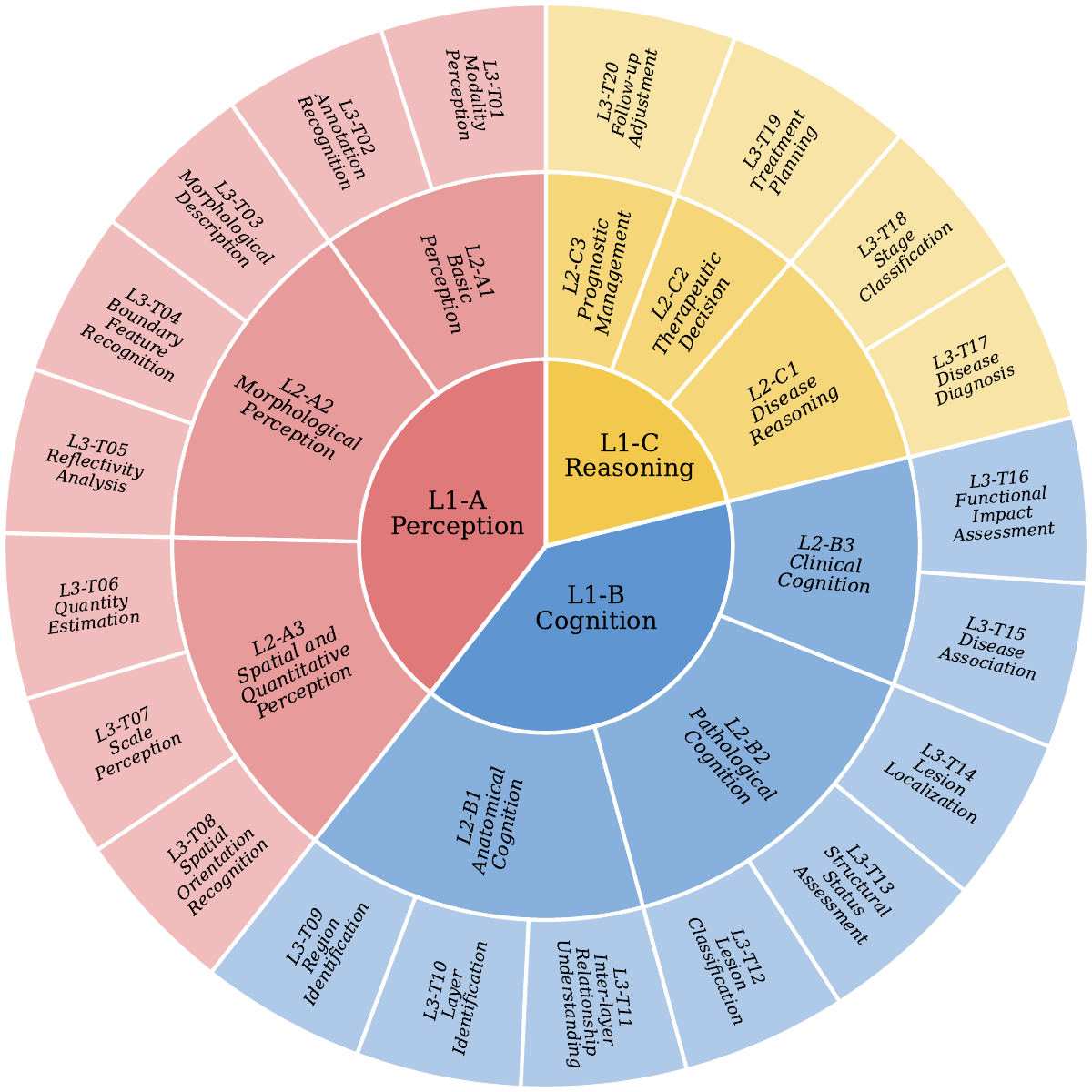}
\caption{Task taxonomy of OCT-Bench.}
\label{fig:fig1}
\end{figure}

In recent years, general-purpose multimodal benchmarks, such as MMBench, MME-RealWorld, MathVista, and SEED-Bench, have substantially advanced the evaluation of multimodal large language models (MLLMs). However, they are primarily designed for natural scenes and cannot effectively assess the understanding of ophthalmic anatomy, retinal lesions, and their clinical significance~\cite{liu2024mmbench,zhang2025mme,lu2024mathvista,peng2025can,jiang2025kore,jiang2026mined,jia2026benchmarking,liu2026general365,jiang2025mmke}. Existing ophthalmology benchmarks mainly focus on medical knowledge, retinal image enhancement, surgical video understanding, or multimodal ophthalmic tasks~\cite{antaki2023evaluating,lim2023benchmarking,zhu2025eyebench,hu2024ophnet}. Nevertheless, a systematic benchmark dedicated to OCT image understanding remains unavailable. As a key imaging modality for retinal disease diagnosis, OCT requires models to recognize fine-grained retinal structures and lesions while integrating medical knowledge for clinical reasoning. To fill this gap, OCT-Bench systematically evaluates MLLMs on OCT image understanding across three progressive levels: visual perception, medical cognition, and clinical reasoning.

\section{OCT-Bench}

\subsection{Hierarchical Capability Taxonomy}
OCT-Bench models OCT understanding as a progressive process from visual perception to medical cognition and clinical reasoning. As shown in Figure~\ref{fig:fig1}, we establish a hierarchical taxonomy with three dimensions, nine capability groups, and 20 fine-grained tasks, following the clinical workflow of OCT interpretation: perceiving visual evidence, linking it to anatomical and pathological concepts, and supporting clinical decision-making.

\subsubsection{Perception}
The Perception dimension evaluates whether a model can accurately extract visual evidence from OCT images, which forms the basis of medical understanding. It covers image attributes, retinal structures, reflectivity patterns, quantitative information, and spatial relationships, assessing the ability of MLLMs to capture fine-grained visual features.

\subsubsection{Cognition}
The Cognition dimension evaluates the transformation of visual information into medical knowledge, focusing on anatomical understanding, pathological recognition, and clinical associations. Models are required to identify retinal structures, localize abnormalities, and link imaging findings with disease states and functional impacts.

\subsubsection{Reasoning}

The Reasoning dimension evaluates whether a model can integrate imaging evidence with medical knowledge for clinical decision-making. It involves disease assessment, treatment planning, and prognosis management, requiring models to synthesize findings and infer appropriate clinical strategies. Separating reasoning from perception and cognition enables OCT-Bench to distinguish visual grounding failures from higher-level inference failures.

\subsection{Benchmark Construction}

\begin{figure*}[ht]
\centering
\includegraphics[width=\textwidth]{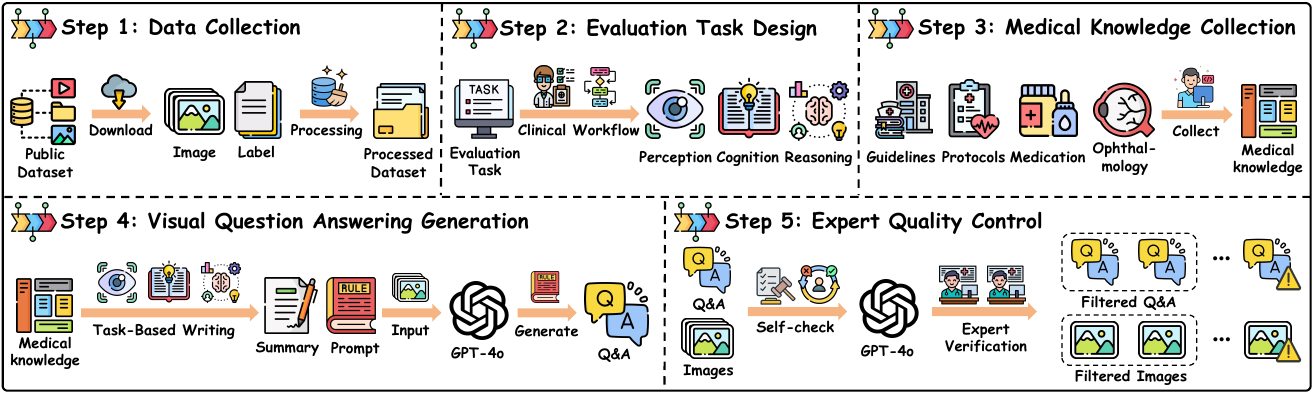}
\caption{Construction pipeline of OCT-Bench. The pipeline comprises data collection, evaluation task design, medical knowledge collection, task-guided VQA generation, and expert quality control.}
\label{fig:fig2}
\end{figure*}

To construct a high-quality benchmark for OCT image understanding, we design a systematic pipeline comprising five stages: data collection, task design, medical knowledge collection, visual question answering generation, and expert quality control, as illustrated in Figure~\ref{fig:fig2}.

\textbf{Step 1: Data Collection.}
We collect OCT images from seven public datasets: OCT5k~\cite{arikan2025oct5k}, OIMHS~\cite{ye2023oimhs}, OCT-C8~\cite{subramanian2022classification}, AMD-SD~\cite{hu2024amd}, OCTDL~\cite{kulyabin2024octdl}, MMC-AMD~\cite{jbhi22-mmcamd}, and GOALS~\cite{fang2022dataset}. We standardize heterogeneous annotations, including categories, bounding boxes, segmentation masks, and clinical attributes. The unified dataset covers diverse diseases, anatomical structures, and lesion information, supporting multi-level OCT understanding.

\textbf{Step 2: Evaluation Task Design.}
Following the clinical OCT interpretation workflow, we organize the evaluation into three capability levels: Perception, Cognition, and Reasoning, which are further divided into nine capability groups and 20 fine-grained tasks. We define each task according to its evaluation objective and knowledge boundary while minimizing overlap among different capabilities.

\textbf{Step 3: Medical Knowledge Collection.}
For cognition and reasoning tasks, we collect medical knowledge from evidence-based guidelines, expert consensus statements, and authoritative ophthalmic references from organizations such as the American Academy of Ophthalmology (AAO) and Chinese Medical Association (CMA), along with OCT reference books~\cite{xun2023evidence,vemulakonda2025age,lim2025diabetic,kim2025idiopathic,kovach2025retinal,duker2021handbook}. These materials are organized into task-specific knowledge constraints to support question construction for disease assessment, treatment decisions, prognosis, and follow-up management.

\textbf{Step 4: Visual Question Answering Generation.}
We design dedicated generation instructions for each task and provide GPT-4o~\cite{hurst2024gpt} with the task description, OCT image, annotation information, and relevant medical knowledge to generate four-option multiple-choice questions. This task-driven strategy aligns each question with its target capability and produces candidate VQA samples covering visual attributes, anatomical structures, lesion characteristics, disease status, therapeutic decisions, and prognostic management.

\textbf{Step 5: Expert Quality Control.}
We adopt a two-stage quality-control strategy. First, GPT-4o automatically checks question quality, answer uniqueness, and image--text consistency. Domain experts then manually review the samples for medical correctness, visual answerability, task relevance, and ambiguity, revising or removing problematic instances. The final OCT-Bench contains 10,076 expert-verified multiple-choice questions.

\subsection{Data Analysis}

We analyze the constructed benchmark from two complementary perspectives: coverage and reliability. In terms of coverage, OCT-Bench spans clinically relevant OCT understanding across 10 disease categories, 2 types of region recognition, 5 types of retinal layer recognition, and 10 types of lesion recognition. We also inspect distributions across tasks, diseases, anatomical regions, retinal layers, and lesion types to reduce label concentration and annotation artifacts. Detailed statistics and distribution analyses are provided in the appendix.

In terms of reliability, we further assess VQA quality before finalizing the benchmark. Each candidate question is checked for image grounding, answer uniqueness, clinical validity, and task alignment, ensuring that the answer is supported by visible OCT evidence or standardized annotations, contains no overlapping distractors, follows authoritative medical references, and matches the intended capability level. Samples with insufficient visual evidence, ambiguity, or language-only cues are revised and rechecked, or removed.

\section{Experiment}

\subsection{Experimental Setup}

We evaluate 20 representative MLLMs on OCT-Bench, covering proprietary, general-purpose, and medical-domain models. Proprietary models include GPT-5.4-mini, Gemini-2.5-flash~\cite{comanici2025gemini}, and Grok-4-fast. General-purpose open-source models include Phi-3-Vision-128K~\cite{bilenko2024new,abdin2024phi}, InternVL2.5 series~\cite{chen2024expanding}, Qwen2.5-VL series~\cite{bai2025qwen2}, Gemma-3-12B~\cite{team2024gemma}, and LLaVA series~\cite{liu2024improved,li2024llava}. Medical-domain models include HealthGPT-M3~\cite{lin2025healthgpt}, MedGemma-4B~\cite{sellergren2025medgemma}, Lingshu series~\cite{xu2025lingshu}, Hulu-Med series~\cite{jiang2025hulu}, MediX-R1-8B~\cite{mullappilly2026medix}, Fleming-VL-8B~\cite{shu2025fleming}, and HealthGPT-Pro-8B. This selection enables comprehensive comparison across general and specialized MLLMs. All models are evaluated under a unified zero-shot setting without fine-tuning or in-context demonstrations.

\begin{table}[!t]
\centering
\resizebox{\columnwidth}{!}{
\begin{tabular}{l|c|c|c|c}
\toprule
\textbf{Model}
& \textbf{Overall}
& \textbf{Perception}
& \textbf{Cognition}
& \textbf{Reasoning} \\
\midrule
\multicolumn{5}{c}{\textbf{Closed-Source Models}}\\
\midrule
\modellogo{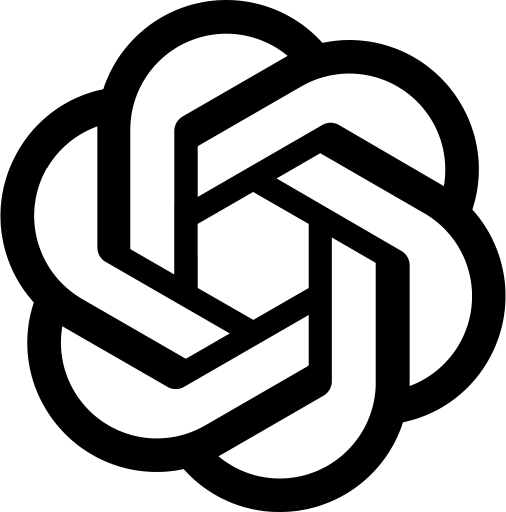}{GPT-5.4-mini}
& \textbf{62.01} & \textbf{75.82} & \underline{59.98} & 38.45 \\
\modellogo{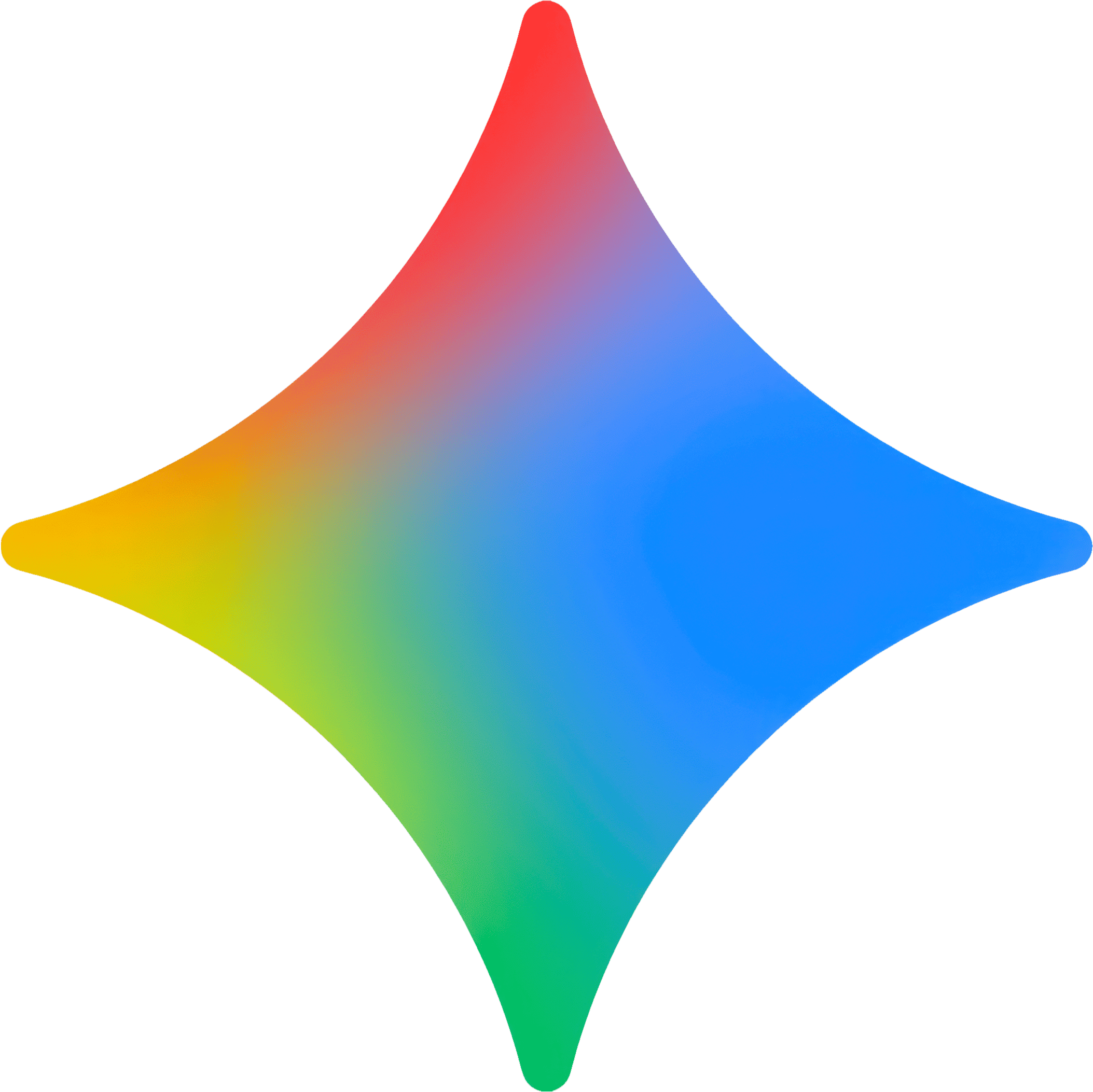}{Gemini-2.5-flash}
& \underline{60.46} & 67.74 & \textbf{64.22} & 38.40 \\
\modellogo{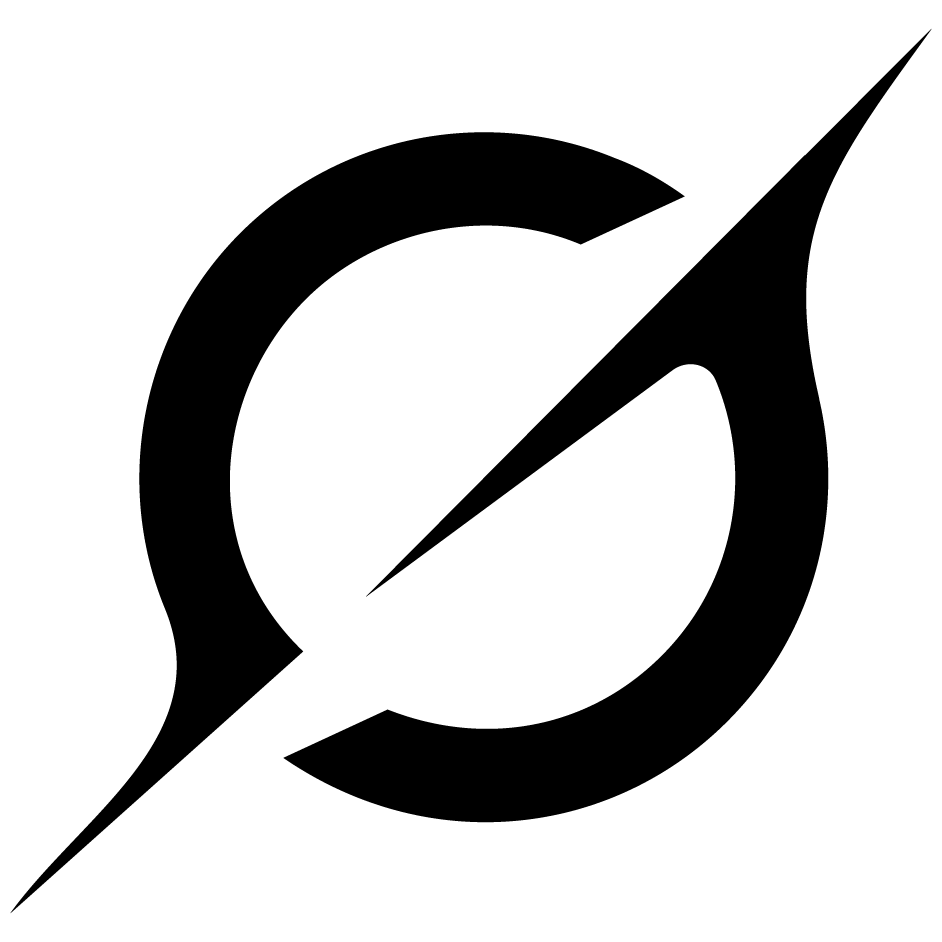}{Grok-4-fast}
& 52.42 & 63.48 & 51.44 & 32.23 \\
\midrule
\multicolumn{5}{c}{\textbf{Open-Source Models}}\\
\midrule
\modellogo{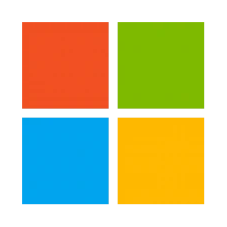}{Phi-3-Vision-128k}
& 42.05 & 60.95 & 32.18 & 23.99 \\
\modellogo{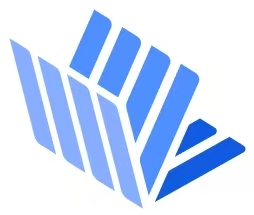}{InternVL2.5-4B}
& 46.62 & 61.19 & 42.29 & 26.17 \\
\modellogo{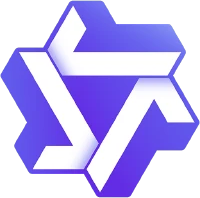}{Qwen2.5-VL-7B}
& 49.49 & 66.74 & 42.66 & 28.65 \\
\modellogo{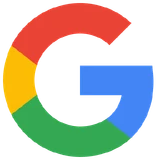}{Gemma-3-12B}
& 44.73 & 62.62 & 36.56 & 25.29 \\
\modellogo{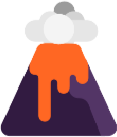}{LLaVA-1.5-13B}
& 39.60 & 54.21 & 28.76 & 32.07 \\
\modellogo{logos/internvl.png}{InternVL2.5-26B}
& 54.54 & \underline{72.46} & 49.36 & 29.05 \\
\modellogo{logos/qwen.png}{Qwen2.5-VL-32B}
& 52.44 & 65.98 & 48.62 & 33.00 \\
\modellogo{logos/llava.png}{LLaVA-1.6-34B}
& 50.53 & 65.31 & 46.22 & 29.60 \\
\midrule
\multicolumn{5}{c}{\textbf{Medical-Domain Models}}\\
\midrule
\modellogo{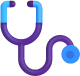}{HealthGPT-M3}
& 45.07 & 61.55 & 38.06 & 26.16 \\
\modellogo{logos/google.png}{MedGemma-4B}
& 42.18 & 55.28 & 35.18 & 29.98 \\
\modellogo{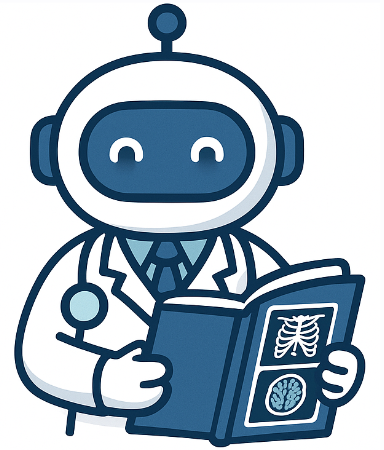}{Lingshu-7B}
& 51.53 & 64.47 & 44.76 & 39.18 \\
\modellogo{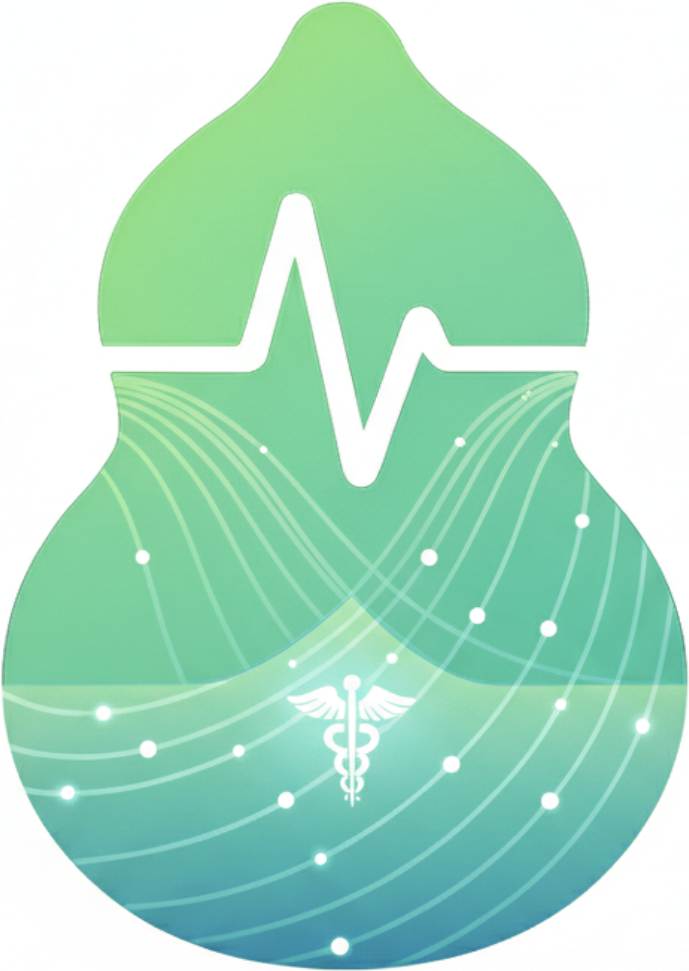}{Hulu-Med-7B}
& 54.13 & 65.97 & 50.25 & 38.22 \\
\modellogo{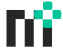}{MediX-R1-8B}
& 51.68 & 65.11 & 48.79 & 30.59 \\
\modellogo{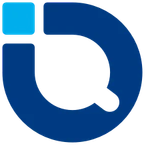}{Fleming-VL-8B}
& 53.62 & 67.84 & 48.76 & 34.93 \\
\modellogo{logos/healthgpt.png}{HealthGPT-Pro-8B}
& 55.58 & 72.40 & 51.61 & 29.88 \\
\modellogo{logos/lingshu.png}{Lingshu-32B}
& 56.58 & 62.53 & 58.75 & \underline{40.35} \\
\modellogo{logos/hulu.png}{Hulu-Med-32B}
& 58.40 & 67.49 & 57.07 & \textbf{42.89} \\
\bottomrule
\end{tabular}
}
\caption{Overall and dimension-average performance on OCT-Bench (\%). The best and second-best results in each column are highlighted in bold and underlined, respectively.}
\label{table2}
\end{table}

\subsection{Evaluation Strategy}

All OCT-Bench tasks are formulated as multiple-choice questions (MCQs) for standardized evaluation. Each question contains four options (A--D) with one correct answer. Given an OCT image and question, models are required to output only the option letter, which is compared with the ground truth. Invalid responses are counted as incorrect. Accuracy is used as the primary metric, enabling objective and fair comparison across tasks and models.

\begin{table*}[t]
\centering
\setlength{\tabcolsep}{3.5pt}
\resizebox{\textwidth}{!}{
\begin{tabular}{l|cc|ccc|ccc|ccc|ccc|cc|cc|c|c}
\toprule
&
\multicolumn{8}{c|}{\textbf{Perception}}
&
\multicolumn{8}{c|}{\textbf{Cognition}}
&
\multicolumn{4}{c}{\textbf{Reasoning}}
\\
\cmidrule(lr){2-9}
\cmidrule(lr){10-17}
\cmidrule(lr){18-21}

\textbf{Model}

&
\multicolumn{2}{c|}{\textbf{A1}}
&
\multicolumn{3}{c|}{\textbf{A2}}
&
\multicolumn{3}{c|}{\textbf{A3}}
&
\multicolumn{3}{c|}{\textbf{B1}}
&
\multicolumn{3}{c|}{\textbf{B2}}
&
\multicolumn{2}{c|}{\textbf{B3}}
&
\multicolumn{2}{c|}{\textbf{C1}}
&
\multicolumn{1}{c|}{\textbf{C2}}
&
\multicolumn{1}{c}{\textbf{C3}}
\\
\cmidrule(lr){2-3}
\cmidrule(lr){4-6}
\cmidrule(lr){7-9}
\cmidrule(lr){10-12}
\cmidrule(lr){13-15}
\cmidrule(lr){16-17}
\cmidrule(lr){18-19}
\cmidrule(lr){20-20}
\cmidrule(lr){21-21}
&
\textbf{T01} & \textbf{T02}
&
\textbf{T03} & \textbf{T04} & \textbf{T05}
&
\textbf{T06} & \textbf{T07} & \textbf{T08}
&
\textbf{T09} & \textbf{T10} & \textbf{T11}
&
\textbf{T12} & \textbf{T13} & \textbf{T14}
&
\textbf{T15} & \textbf{T16}
&
\textbf{T17} & \textbf{T18}
&
\textbf{T19}
&
\textbf{T20}
\\
\midrule
\multicolumn{21}{c}{\textbf{Closed-Source Models}} \\
\midrule
\modellogo{logos/openai.png}{GPT-5.4-mini} &
\textbf{100.00} & 98.38 & 43.31 & 69.94 & \textbf{61.10} & \textbf{90.70} & \textbf{67.05} & \textbf{76.10} &
61.17 & \textbf{56.79} & 46.89 & 35.01 & \textbf{81.18} & \textbf{92.06} & 34.66 & \underline{72.09} &
38.42 & 30.68 & 40.87 & 43.83 \\
\modellogo{logos/gemini.png}{Gemini-2.5-flash} &
\textbf{100.00} & \underline{99.54} & 32.56 & 60.13 & 57.96 & 87.79 & 49.42 & 54.52 &
\underline{92.42} & 37.26 & 44.69 & 37.87 & 78.71 & 85.98 & \underline{52.15} & \textbf{84.65} &
19.31 & \textbf{34.00} & 49.36 & \textbf{50.94} \\
\modellogo{logos/grok.png}{Grok-4-fast} &
\textbf{100.00} & 95.36 & 33.14 & 73.31 & 54.22 & 62.50 & 42.46 & 46.87 &
51.89 & 30.53 & 35.62 & 37.73 & 66.92 & 81.54 & \textbf{58.01} & 49.30 &
23.75 & \textbf{34.00} & 29.56 & 41.62 \\
\midrule
\multicolumn{21}{c}{\textbf{Open-Source Models}} \\
\midrule
\modellogo{logos/microsoft.png}{Phi-3-Vision-128k} &
95.41 & 97.91 & 9.01 & 74.28 & 36.15 & 54.65 & 60.79 & 59.40 &
40.91 & 23.03 & 36.01 & 28.89 & 62.36 & 12.15 & 45.97 & 8.14 &
17.18 & 32.45 & 19.79 & 26.54 \\
\modellogo{logos/internvl.png}{InternVL2.5-4B} &
\textbf{100.00} & 97.45 & 24.13 & 55.95 & 38.51 & 77.62 & 54.06 & 41.76 &
70.45 & 30.27 & 32.64 & 36.89 & 50.19 & 26.17 & 47.75 & 43.95 &
24.52 & 33.11 & 21.85 & 25.20 \\
\modellogo{logos/qwen.png}{Qwen2.5-VL-7B} &
98.06 & \textbf{99.77} & 43.31 & 43.09 & 46.56 & \underline{90.12} & 44.08 & 68.91 &
44.89 & 27.68 & 34.97 & 31.84 & 58.75 & 63.55 & 42.62 & 36.98 &
23.36 & 32.23 & 30.33 & 28.69 \\
\modellogo{logos/google.png}{Gemma-3-12B} &
99.47 & \textbf{99.77} & 25.58 & 81.35 & 48.33 & 58.72 & 40.14 & 47.56 &
51.70 & 25.49 & 29.15 & 34.22 & 40.87 & 42.99 & 31.10 & 36.98 &
24.13 & 25.83 & 25.71 & 25.47 \\
\modellogo{logos/llava.png}{LLaVA-1.5-13B} &
68.25 & 93.27 & 11.34 & \underline{81.67} & 36.35 & 62.79 & 56.61 & 23.43 &
10.80 & 20.44 & 23.19 & 30.15 & 35.17 & 28.50 & 40.63 & 41.16 &
25.68 & 33.11 & 36.25 & 33.24 \\
\modellogo{logos/internvl.png}{InternVL2.5-26B} &
\textbf{100.00} & 99.30 & \textbf{45.35} & 76.85 & 39.49 & 86.92 & 61.02 & \underline{70.77} &
63.07 & 28.98 & 24.87 & 32.96 & 69.39 & 74.53 & 47.12 & 53.95 &
24.52 & 31.35 & 37.28 & 23.06 \\
\modellogo{logos/qwen.png}{Qwen2.5-VL-32B} &
99.12 & 99.30 & 30.81 & 41.48 & 52.85 & \underline{90.12} & 45.48 & 68.68 &
65.15 & 21.99 & 33.16 & 36.89 & 60.46 & 68.69 & \underline{52.15} & 50.47 &
21.62 & 33.11 & 36.25 & 41.02 \\
\modellogo{logos/llava.png}{LLaVA-1.6-34B} &
\textbf{100.00} & 97.22 & 35.17 & \underline{81.67} & 36.54 & 47.97 & \textbf{67.05} & 56.84 &
50.38 & 23.80 & 34.97 & 25.11 & 62.93 & \underline{87.62} & 47.96 & 36.98 &
32.05 & 28.26 & 27.25 & 30.83 \\
\midrule
\multicolumn{21}{c}{\textbf{Medical-Domain Models}} \\
\midrule
\modellogo{logos/healthgpt.png}{HealthGPT-M3} &
\textbf{100.00} & 93.04 & 0.87 & \textbf{82.32} & \textbf{61.10} & 75.00 & 64.73 & 15.31 &
72.73 & 32.21 & 38.99 & 23.28 & 11.79 & 33.64 & 43.46 & 48.37 &
32.63 & 33.11 & 18.51 & 20.38 \\
\modellogo{logos/google.png}{MedGemma-4B} &
\underline{99.65} & 83.99 & 7.27 & 74.60 & 33.01 & 80.81 & 38.05 & 24.83 &
53.98 & 24.45 & 31.09 & \underline{39.55} & 5.13 & 41.36 & 44.92 & 40.93 &
39.77 & 33.11 & 16.20 & 30.83 \\
\modellogo{logos/lingshu.png}{Lingshu-7B} &
99.29 & 99.30 & 2.62 & 70.42 & 50.49 & \underline{90.12} & 51.97 & 51.51 &
61.17 & 29.11 & 31.09 & \textbf{39.97} & 51.90 & 67.76 & 28.48 & 48.60 &
30.31 & 28.48 & 48.33 & \underline{49.60} \\
\modellogo{logos/hulu.png}{Hulu-Med-7B} &
\textbf{100.00} & \underline{99.54} & 5.81 & 80.39 & 59.92 & 84.88 & 46.64 & 50.58 &
89.58 & 28.33 & 30.31 & 30.72 & 64.45 & 70.09 & 39.69 & 48.84 &
\underline{44.59} & 32.89 & 42.42 & 32.98 \\
\modellogo{logos/medix.png}{MediX-R1-8B} &
\textbf{100.00} & 99.30 & 33.14 & 81.35 & 46.17 & 63.95 & 55.92 & 41.07 &
50.95 & 29.11 & 39.12 & 27.63 & \underline{79.66} & 62.85 & 48.69 & 52.33 &
26.83 & \underline{33.33} & 19.02 & 43.16 \\
\modellogo{logos/fleming.png}{Fleming-VL-8B} &
\textbf{100.00} & 99.30 & \underline{44.19} & 30.87 & 55.99 & 86.34 & 64.04 & 61.95 &
69.13 & 31.95 & 33.03 & 39.41 & 64.45 & 70.09 & 38.74 & 43.26 &
41.12 & 32.67 & 38.56 & 27.35 \\
\modellogo{logos/healthgpt.png}{HealthGPT-Pro-8B} &
\textbf{100.00} & \textbf{99.77} & 21.80 & 78.46 & 59.14 & 87.50 & \underline{66.36} & 66.13 &
91.10 & 36.87 & 43.39 & 31.42 & 51.14 & 63.79 & 45.86 & 49.30 &
30.89 & 32.01 & 29.82 & 26.81 \\
\modellogo{logos/lingshu.png}{Lingshu-32B} &
98.94 & \textbf{99.77} & 8.43 & 74.28 & 31.24 & \underline{90.12} & 30.63 & 66.82 &
89.96 & \underline{43.60} & \textbf{59.20} & 38.99 & 61.98 & 73.36 & 45.45 & 57.44 &
34.36 & 32.67 & \textbf{56.04} & 38.34 \\
\modellogo{logos/hulu.png}{Hulu-Med-32B} &
\underline{99.65} & \textbf{99.77} & 9.30 & 80.71 & \underline{60.12} & 85.47 & 60.79 & 44.08 &
\textbf{97.35} & 36.48 & \underline{47.28} & 34.22 & 60.84 & 76.40 & 50.26 & 53.72 &
\textbf{47.68} & 32.89 & \underline{53.47} & 37.53 \\
\bottomrule
\end{tabular}
}
\caption{Performance of different models on tasks T01--T20 (\%).}
\label{table3}
\end{table*}

\subsection{Main Results}

Table~\ref{table2} summarizes the overall performance and average results across the three capability dimensions. Four key observations emerge from these results.

\textbf{Current MLLMs remain far from reliable OCT understanding.}
GPT-5.4-mini achieves the best overall accuracy (62.0\%), followed by Gemini-2.5-flash (60.5\%) and Hulu-Med-32B (58.4\%). Although all models outperform the 25\% random-guess baseline, even the best model answers nearly 38\% of questions incorrectly, indicating that reliable clinical OCT interpretation remains out of reach. Moreover, no model consistently leads across all capability dimensions: GPT-5.4-mini performs best on Perception (75.8\%), Gemini-2.5-flash on Cognition (64.2\%), and Hulu-Med-32B on Reasoning (42.9\%). This divergence shows that overall accuracy alone can mask substantial differences in capability profiles.

\textbf{Performance degrades as capability requirements progress from perception to reasoning.}
Using the best result in each dimension, accuracy declines from 75.8\% on Perception to 64.2\% on Cognition and 42.9\% on Reasoning, a drop of 32.9 points from the first to the final stage. This trend is also evident within individual models: strong visual perception does not necessarily translate into clinical reasoning. For example, GPT-5.4-mini achieves the highest Perception score but reaches only 38.5\% on Reasoning. Even the best Reasoning result remains below 50\% and only 17.9 points above chance, suggesting that visual evidence alone is insufficient without robust medical knowledge and image-grounded reasoning.



\textbf{Model specialization does not ensure uniform superiority.}
Closed-source models are competitive overall, with GPT-5.4-mini and Gemini-2.5-flash ranking first and second. However, the best performance remains distributed across model types: Gemini-2.5-flash leads Cognition, while a medical-domain model leads Reasoning. Medical models also vary substantially. Although Hulu-Med-32B and Lingshu-32B rank third and fourth overall, HealthGPT-M3 and MedGemma-4B perform worse than several general-purpose models. These results suggest that domain specialization benefits specific capabilities but does not guarantee comprehensive OCT understanding.

\textbf{Scaling produces uneven gains across capability levels.}
Scaling consistently improves overall accuracy within the InternVL, Qwen, Lingshu, and Hulu-Med families, but the gains are uneven across capability levels. For example, scaling Lingshu from 7B to 32B increases Cognition by 14 points, while Perception decreases slightly and Reasoning improves by only 1.2 points. Similar trends are observed for Qwen2.5-VL. These results indicate that scaling strengthens specific capabilities but does not resolve the bottleneck in clinically grounded reasoning, highlighting the need for better visual grounding, image--knowledge alignment, and multi-step clinical inference.

\subsection{Fine-grained Analysis}

Table~\ref{table3} further decomposes model performance into 20 fine-grained tasks, revealing that the difficulty of OCT understanding is highly task-dependent even within the same capability level.

\textbf{Low-level recognition is nearly saturated, but fine-grained visual description remains fragile.}
Most models perform well on modality perception (T01) and annotation recognition (T02), achieving average accuracies of 97.9\% and 97.6\%, demonstrating reliable recognition of imaging attributes and explicit annotations. However, performance drops sharply for subtle morphological description, averaging only 23.4\% on T03 with a best score of 45.4\%. This suggests that MLLMs can recognize OCT images but lack precise representations of retinal shape, deformation, and lesion morphology. Similar limitations appear in scale perception (T07, 53.4\%) and spatial orientation recognition (T08, 51.9\%), indicating insufficient geometric grounding.

\begin{figure}[ht]
\centering
\includegraphics[width=\columnwidth]{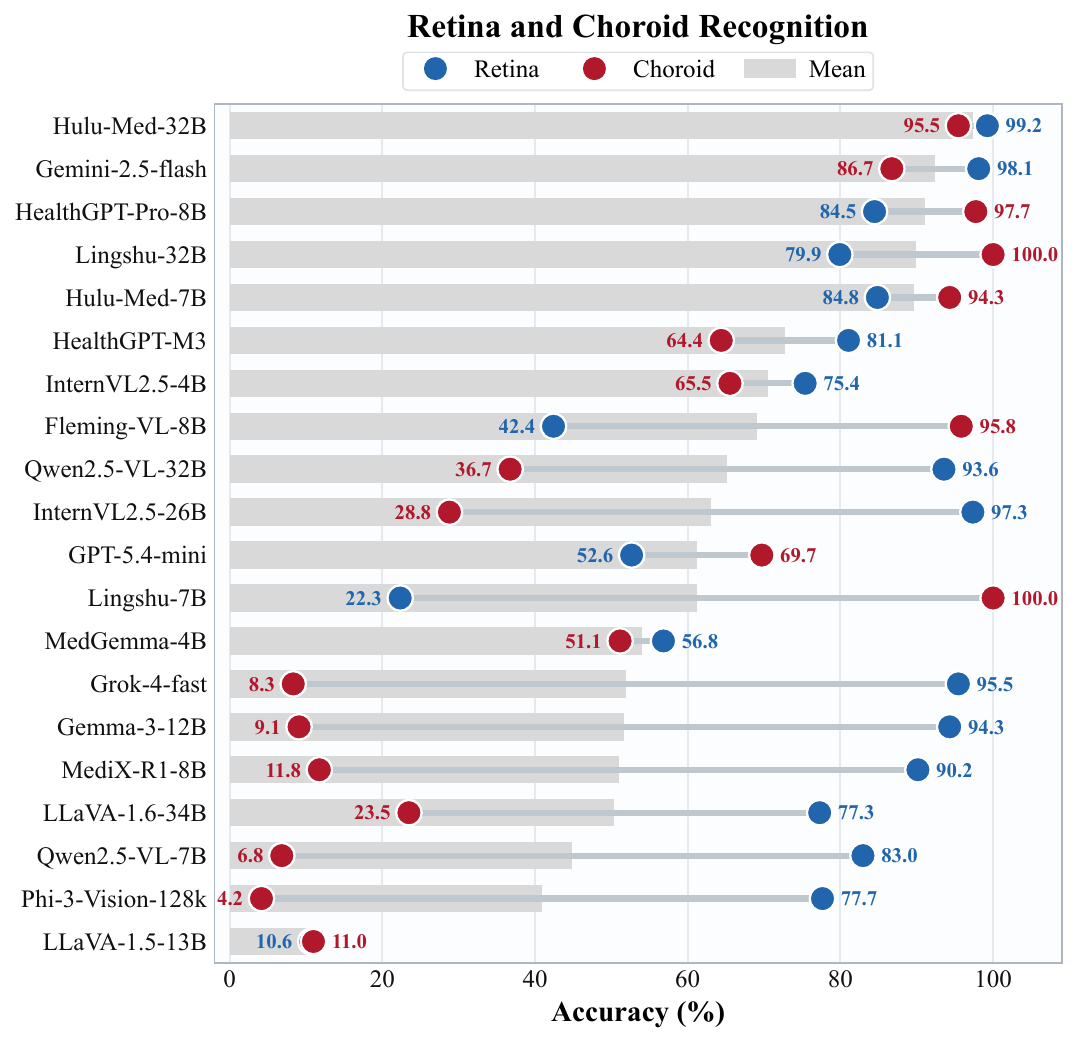}
\caption{Retina and choroid recognition accuracy for task T09 across different models.}
\label{fig:fig4}
\end{figure}

\textbf{Anatomical cognition exposes strong structure-specific bias.}
Although region identification (T09) achieves a best score of 97.4\%, Figure~\ref{fig:fig4} reveals large variation across regions. Retina recognition averages 76.5\%, whereas choroid recognition reaches only 50.7\%. Several models exceed 80\% on retina recognition but fall below 21\% on choroid recognition, while Lingshu-7B and Fleming-VL-8B show the opposite trend. This suggests that high anatomical scores often rely on region-specific bias rather than balanced understanding. The limitation is more pronounced in layer identification (T10), where the average accuracy is only 30.9\% and the best result is 56.8\%, highlighting the difficulty of distinguishing fine retinal layers.

\begin{figure}[ht]
\centering
\includegraphics[width=\columnwidth]{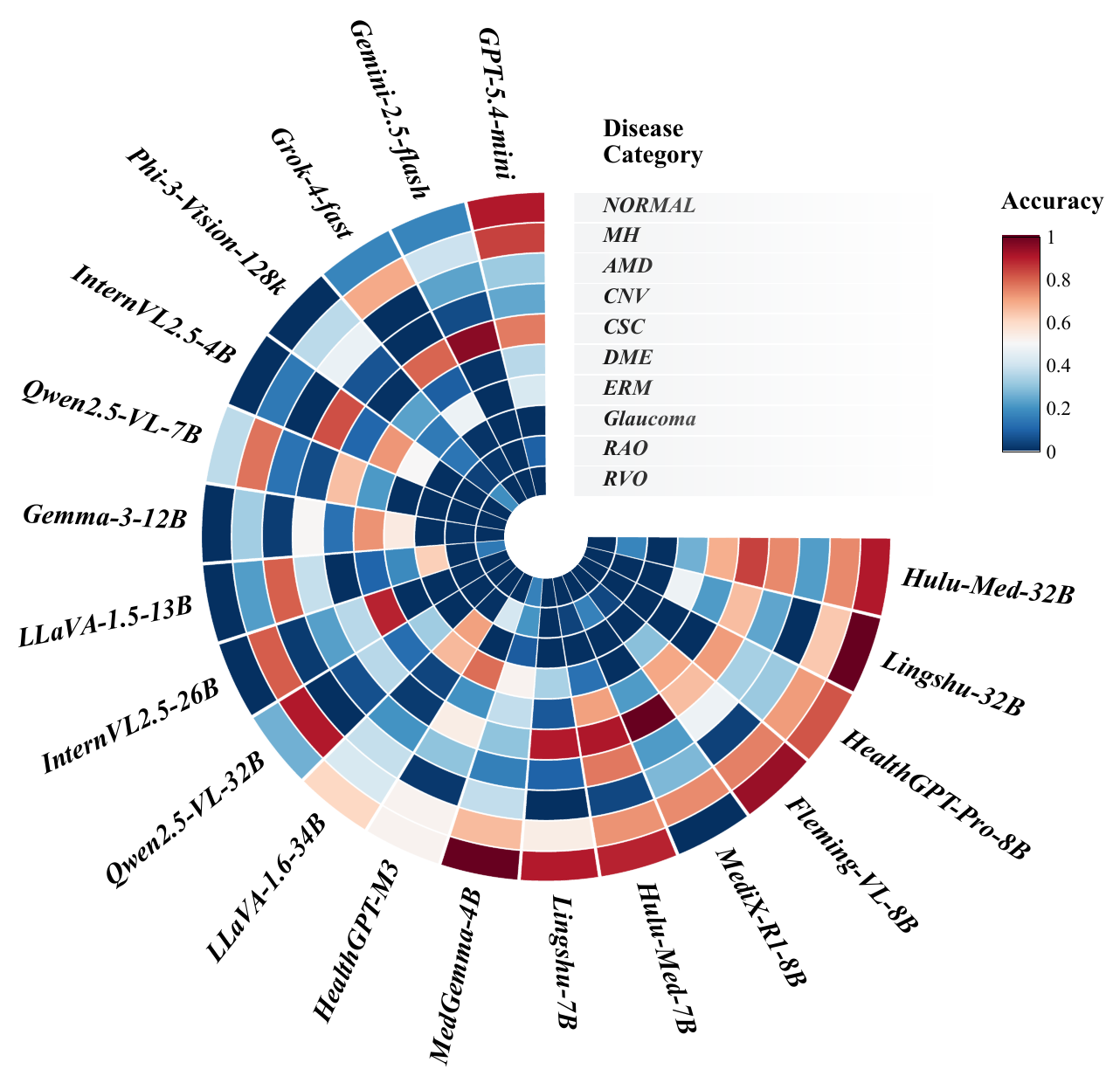}
\caption{Disease-category-level diagnosis accuracy for task T17. }
\label{fig:fig5}
\end{figure}

\textbf{Disease reasoning is dominated by category-dependent shortcuts.}
Disease diagnosis (T17) is one of the most clinically important tasks, yet the average accuracy is only 30.1\% and the best result is 47.7\%. Figure~\ref{fig:fig5} reveals substantial variation across disease categories. Models are relatively more successful on MH and CSC, whose average accuracies are 58.4\% and 53.0\%, but perform poorly on AMD (16.74\%), Glaucoma (7.8\%), RAO (5.5\%), and RVO (2.5\%). Such category imbalance suggests that models tend to rely on salient or frequently observed patterns, while failing on diseases that require subtle vascular, glaucomatous, or differential diagnostic reasoning.

\textbf{Clinical reasoning remains weak even when visual cues are available.}
The reasoning tasks show consistently low ceilings. Stage classification (T18) is tightly clustered between 25.8\% and 34.0\%, indicating that nearly all models struggle to map OCT findings to disease severity. Treatment planning (T19) and follow-up adjustment (T20) achieve best scores of 56.0\% and 50.9\%, respectively, but their average accuracies remain only 33.8\% and 33.9\%. These results imply that models may occasionally recover common clinical priors, yet they cannot reliably integrate anatomical evidence, lesion status, and disease context into coherent decisions.

\begin{figure}[!ht]
\centering
\includegraphics[width=\columnwidth]{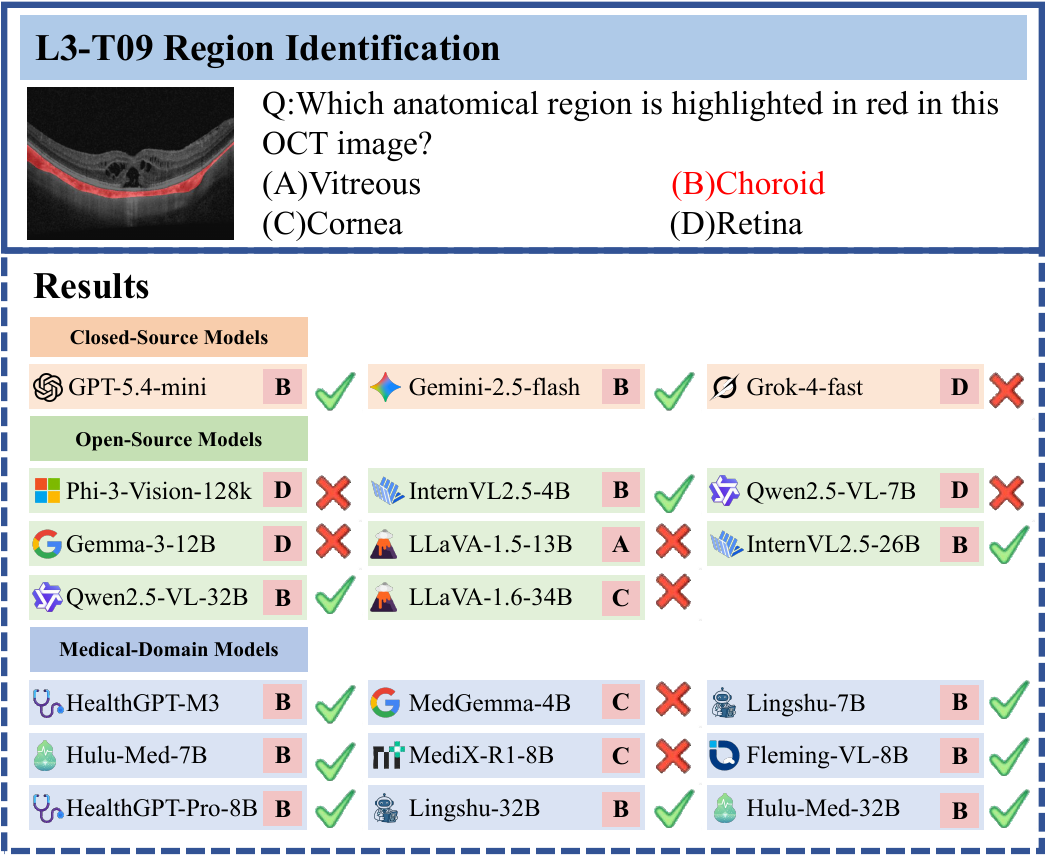}
\caption{Case study of Region Identification.}
\label{fig:fig6}
\end{figure}

\section{Case Study}

To better understand model behavior beyond aggregate scores, Figure~\ref{fig:fig6} presents a representative example from the Region Identification task. The highlighted region lies beneath the retinal layers and corresponds to the choroid. GPT-5.4-mini and Gemini-2.5-flash answer correctly, whereas Grok-4-fast misidentifies it as the retina. General-purpose open-source models show greater variation, frequently confusing the choroid with the retina, vitreous, or cornea. Most medical-domain models correctly recognize the region, although MedGemma-4B and MediX-R1-8B identify it as the cornea. Since the target is explicitly highlighted, these errors primarily reflect insufficient anatomical understanding rather than failed localization. Such misidentification provides an incorrect anatomical premise that may further impair downstream clinical reasoning.


\section{Conclusion}


In this paper, we introduced \textbf{OCT-Bench}, a benchmark for evaluating multimodal large language models on OCT image understanding. Following the clinical interpretation workflow, OCT-Bench organizes OCT understanding into three dimensions: Perception, Cognition, and Reasoning. Built from 4,137 OCT images across seven public datasets, it contains 10,076 expert-verified multiple-choice questions spanning 20 fine-grained tasks, enabling comprehensive evaluation of overall performance and capability bottlenecks.

Evaluations of 20 representative MLLMs reveal that reliable OCT understanding remains challenging. The best model achieves only 62.0\% overall accuracy, with performance dropping markedly from perception to reasoning. Models consistently struggle with fine-grained retinal structures, disease diagnosis, and image-grounded clinical reasoning, while larger model scale and medical-domain adaptation provide limited gains. We hope OCT-Bench will facilitate the development of MLLMs with stronger visual grounding and more reliable ophthalmic reasoning.

\bibliography{aaai2027}

@article{hurst2024gpt,
  title={Gpt-4o system card},
  author={Hurst, Aaron and Lerer, Adam and Goucher, Adam P and Perelman, Adam and Ramesh, Aditya and Clark, Aidan and Ostrow, AJ and Welihinda, Akila and Hayes, Alan and Radford, Alec and others},
  journal={arXiv preprint arXiv:2410.21276},
  year={2024}
}

@article{comanici2025gemini,
  title={Gemini 2.5: Pushing the frontier with advanced reasoning, multimodality, long context, and next generation agentic capabilities},
  author={Comanici, Gheorghe and Bieber, Eric and Schaekermann, Mike and Pasupat, Ice and Sachdeva, Noveen and Dhillon, Inderjit and Blistein, Marcel and Ram, Ori and Zhang, Dan and Rosen, Evan and others},
  journal={arXiv preprint arXiv:2507.06261},
  year={2025}
}

@inproceedings{liu2024improved,
  title={Improved baselines with visual instruction tuning},
  author={Liu, Haotian and Li, Chunyuan and Li, Yuheng and Lee, Yong Jae},
  booktitle={Proceedings of the IEEE/CVF conference on computer vision and pattern recognition},
  pages={26296--26306},
  year={2024}
}

@article{li2024llava,
  title={LLaVA-NeXT-Interleave: Tackling Multi-image, Video, and 3D in Large Multimodal Models},
  author={Li, Feng and Zhang, Renrui and Zhang, Hao and Zhang, Yuanhan and Li, Bo and Li, Wei and Ma, Zejun and Li, Chunyuan},
  journal={arXiv preprint arXiv:2407.07895},
  year={2024}
}

@article{bilenko2024new,
  title={New models added to the Phi-3 family, available on Microsoft Azure},
  author={Bilenko, Misha},
  journal={Microsoft GenAI},
  year={2024}
}

@article{abdin2024phi,
  title={Phi-4 technical report},
  author={Abdin, Marah and Aneja, Jyoti and Behl, Harkirat and Bubeck, S{\'e}bastien and Eldan, Ronen and Gunasekar, Suriya and Harrison, Michael and Hewett, Russell J and Javaheripi, Mojan and Kauffmann, Piero and others},
  journal={arXiv preprint arXiv:2412.08905},
  year={2024}
}

@article{chen2024expanding,
  title={Expanding performance boundaries of open-source multimodal models with model, data, and test-time scaling},
  author={Chen, Zhe and Wang, Weiyun and Cao, Yue and Liu, Yangzhou and Gao, Zhangwei and Cui, Erfei and Zhu, Jinguo and Ye, Shenglong and Tian, Hao and Liu, Zhaoyang and others},
  journal={arXiv preprint arXiv:2412.05271},
  year={2024}
}

@article{bai2025qwen2,
  title={Qwen2. 5-vl technical report},
  author={Bai, Shuai and Chen, Keqin and Liu, Xuejing and Wang, Jialin and Ge, Wenbin and Song, Sibo and Dang, Kai and Wang, Peng and Wang, Shijie and Tang, Jun and others},
  journal={arXiv preprint arXiv:2502.13923},
  year={2025}
}

@article{team2024gemma,
  title={Gemma: Open models based on gemini research and technology},
  author={Team, Gemma and Mesnard, Thomas and Hardin, Cassidy and Dadashi, Robert and Bhupatiraju, Surya and Pathak, Shreya and Sifre, Laurent and Rivi{\`e}re, Morgane and Kale, Mihir Sanjay and Love, Juliette and others},
  journal={arXiv preprint arXiv:2403.08295},
  year={2024}
}

@article{lin2025healthgpt,
  title={Healthgpt: A medical large vision-language model for unifying comprehension and generation via heterogeneous knowledge adaptation},
  author={Lin, Tianwei and Zhang, Wenqiao and Li, Sijing and Yuan, Yuqian and Yu, Binhe and Li, Haoyuan and He, Wanggui and Jiang, Hao and Li, Mengze and Song, Xiaohui and others},
  journal={arXiv preprint arXiv:2502.09838},
  year={2025}
}

@article{sellergren2025medgemma,
  title={Medgemma technical report},
  author={Sellergren, Andrew and Kazemzadeh, Sahar and Jaroensri, Tiam and Kiraly, Atilla and Traverse, Madeleine and Kohlberger, Timo and Xu, Shawn and Jamil, Fayaz and Hughes, C{\'\i}an and Lau, Charles and others},
  journal={arXiv preprint arXiv:2507.05201},
  year={2025}
}

@article{xu2025lingshu,
  title={Lingshu: A generalist foundation model for unified multimodal medical understanding and reasoning},
  author={Xu, Weiwen and Chan, Hou Pong and Li, Long and Aljunied, Mahani and Yuan, Ruifeng and Wang, Jianyu and Xiao, Chenghao and Chen, Guizhen and Liu, Chaoqun and Li, Zhaodonghui and others},
  journal={arXiv preprint arXiv:2506.07044},
  year={2025}
}

@article{jiang2025hulu,
  title={Hulu-med: A transparent generalist model towards holistic medical vision-language understanding},
  author={Jiang, Songtao and Wang, Yuan and Song, Sibo and Hu, Tianxiang and Zhou, Chenyi and Pu, Bin and Zhang, Yan and Yang, Zhibo and Feng, Yang and Zhou, Joey Tianyi and others},
  journal={arXiv preprint arXiv:2510.08668},
  year={2025}
}

@article{mullappilly2026medix,
  title={Medix-r1: Open ended medical reinforcement learning},
  author={Mullappilly, Sahal Shaji and Kurpath, Mohammed Irfan and Mohamed, Omair and Zidan, Mohamed and Khan, Fahad and Khan, Salman and Anwer, Rao and Cholakkal, Hisham},
  journal={arXiv preprint arXiv:2602.23363},
  year={2026}
}

@article{shu2025fleming,
  title={Fleming-VL: Towards Universal Medical Visual Reasoning with Multimodal LLMs},
  author={Shu, Yan and Liu, Chi and Chen, Robin and Li, Derek and Dai, Bryan},
  journal={arXiv preprint arXiv:2511.00916},
  year={2025}
}

@inproceedings{liu2024mmbench,
  title={Mmbench: Is your multi-modal model an all-around player?},
  author={Liu, Yuan and Duan, Haodong and Zhang, Yuanhan and Li, Bo and Zhang, Songyang and Zhao, Wangbo and Yuan, Yike and Wang, Jiaqi and He, Conghui and Liu, Ziwei and others},
  booktitle={European conference on computer vision},
  pages={216--233},
  year={2024},
  organization={Springer}
}

@inproceedings{zhang2025mme,
  title={Mme-realworld: Could your multimodal llm challenge high-resolution real-world scenarios that are difficult for humans?},
  author={Zhang, YiFan and Zhang, Huanyu and Tian, Haochen and Fu, Chaoyou and Zhang, Shuangqing and Wu, Junfei and Li, Feng and Wang, Kun and Wen, Qingsong and Zhang, Zhang and others},
  booktitle={International Conference on Learning Representations},
  volume={2025},
  pages={89655--89701},
  year={2025}
}

@inproceedings{lu2024mathvista,
  title={Mathvista: Evaluating mathematical reasoning of foundation models in visual contexts},
  author={Lu, Pan and Bansal, Hritik and Xia, Tony and Liu, Jiacheng and Li, Chunyuan and Hajishirzi, Hannaneh and Cheng, Hao and Chang, Kai-Wei and Galley, Michel and Gao, Jianfeng},
  booktitle={International Conference on Learning Representations},
  volume={2024},
  pages={23439--23554},
  year={2024}
}

@article{antaki2023evaluating,
  title={Evaluating the performance of ChatGPT in ophthalmology: an analysis of its successes and shortcomings},
  author={Antaki, Fares and Touma, Samir and Milad, Daniel and El-Khoury, Jonathan and Duval, Renaud},
  journal={Ophthalmology science},
  volume={3},
  number={4},
  pages={100324},
  year={2023},
  publisher={Elsevier}
}

@article{lim2023benchmarking,
  title={Benchmarking large language models’ performances for myopia care: a comparative analysis of ChatGPT-3.5, ChatGPT-4.0, and Google Bard},
  author={Lim, Zhi Wei and Pushpanathan, Krithi and Yew, Samantha Min Er and Lai, Yien and Sun, Chen-Hsin and Lam, Janice Sing Harn and Chen, David Ziyou and Goh, Jocelyn Hui Lin and Tan, Marcus Chun Jin and Sheng, Bin and others},
  journal={EBioMedicine},
  volume={95},
  year={2023},
  publisher={Elsevier}
}

@article{zhu2025eyebench,
  title={Eyebench: A call for more rigorous evaluation of retinal image enhancement},
  author={Zhu, Wenhui and Dong, Xuanzhao and Li, Xin and Xiong, Yujian and Chen, Xiwen and Qiu, Peijie and Vasa, Vamsi Krishna and Yang, Zhangsihao and Su, Yi and Dumitrascu, Oana and others},
  journal={arXiv preprint arXiv:2502.14260},
  year={2025}
}

@inproceedings{hu2024ophnet,
  title={Ophnet: A large-scale video benchmark for ophthalmic surgical workflow understanding},
  author={Hu, Ming and Xia, Peng and Wang, Lin and Yan, Siyuan and Tang, Feilong and Xu, Zhongxing and Luo, Yimin and Song, Kaimin and Leitner, Jurgen and Cheng, Xuelian and others},
  booktitle={European Conference on Computer Vision},
  pages={481--500},
  year={2024},
  organization={Springer}
}

@inproceedings{qin2025lmod,
  title={Lmod: A large multimodal ophthalmology dataset and benchmark for large vision-language models},
  author={Qin, Zhenyue and Yin, Yu and Campbell, Dylan and Wu, Xuansheng and Zou, Ke and Liu, Ninghao and Tham, Yih Chung and Zhang, Xiuzhen Jenny and Chen, Qingyu},
  booktitle={Findings of the Association for Computational Linguistics: NAACL 2025},
  pages={2501--2522},
  year={2025}
}

@article{fu2026mmku,
  title={Mmku-bench: A multimodal update benchmark for diverse visual knowledge},
  author={Fu, Baochen and Du, Yuntao and Chang, Cheng and Jin, Baihao and Deng, Wenzhi and Xu, Muhao and Yan, Hongmei and Song, Weiye and Wan, Yi},
  journal={arXiv preprint arXiv:2603.15117},
  year={2026}
}

@article{huang1991optical,
  title={Optical coherence tomography},
  author={Huang, David and Swanson, Eric A and Lin, Charles P and Schuman, Joel S and Stinson, William G and Chang, Warren and Hee, Michael R and Flotte, Thomas and Gregory, Kenton and Puliafito, Carmen A and others},
  journal={science},
  volume={254},
  number={5035},
  pages={1178--1181},
  year={1991},
  publisher={American Association for the Advancement of Science}
}

@article{wu2024visionllm,
  title={Visionllm v2: An end-to-end generalist multimodal large language model for hundreds of vision-language tasks},
  author={Wu, Jiannan and Zhong, Muyan and Xing, Sen and Lai, Zeqiang and Liu, Zhaoyang and Chen, Zhe and Wang, Wenhai and Zhu, Xizhou and Lu, Lewei and Lu, Tong and others},
  journal={Advances in Neural Information Processing Systems},
  volume={37},
  pages={69925--69975},
  year={2024}
}

@article{caffagni2024revolution,
  title={The revolution of multimodal large language models: A survey},
  author={Caffagni, Davide and Cocchi, Federico and Barsellotti, Luca and Moratelli, Nicholas and Sarto, Sara and Baraldi, Lorenzo and Cornia, Marcella and Cucchiara, Rita},
  journal={Findings of the association for computational linguistics: ACL 2024},
  pages={13590--13618},
  year={2024}
}

@article{bai2025label,
  title={Label-semantic-based prompt tuning for vision transformer adaptation in medical image analysis},
  author={Bai, Yu and Bai, Liang and Yang, Xian and Liang, Jiye},
  journal={IEEE Transactions on Circuits and Systems for Video Technology},
  year={2025},
  publisher={IEEE}
}

@article{lin2025taming,
  title={Taming vision-language models for medical image analysis: A comprehensive review},
  author={Lin, Haoneng and Xu, Cheng and Qin, Jing},
  journal={arXiv preprint arXiv:2506.18378},
  year={2025}
}

@article{fang2025research,
  title={Research progress on AI-assisted screening and prediction of systemic diseases based on retinal images},
  author={Fang, Pinqi and Wu, Yiting and He, Yufeng and Li, Haoxuan and Guan, Zhouyu and Wang, Xiangning and Chen, Tingli and Shen, Jie},
  journal={The Visual Computer},
  volume={41},
  number={12},
  pages={9509--9537},
  year={2025},
  publisher={Springer}
}

@article{gurumoorthy2025role,
  title={The role of artificial intelligence in monitoring glaucoma progression using optical coherence tomography},
  author={Gurumoorthy, Sushmitha and Nayaki, NV Meena Lochana and Vendhan, K Ezhil},
  journal={Indian Journal of Clinical and Experimental Ophthalmology},
  volume={11},
  number={4},
  pages={631--640},
  year={2025}
}

@article{wang2024advances,
  title={Advances and prospects of multi-modal ophthalmic artificial intelligence based on deep learning: a review},
  author={Wang, Shaopan and He, Xin and Jian, Zhongquan and Li, Jie and Xu, Changsheng and Chen, Yuguang and Liu, Yuwen and Chen, Han and Huang, Caihong and Hu, Jiaoyue and others},
  journal={Eye and Vision},
  volume={11},
  number={1},
  pages={38},
  year={2024},
  publisher={Springer}
}

@article{chen2024visual,
  title={Visual Question Answering in Ophthalmology: A Progressive and Practical Perspective},
  author={Chen, Xiaolan and Chen, Ruoyu and Xu, Pusheng and Zhang, Weiyi and Shang, Xianwen and He, Mingguang and Shi, Danli},
  journal={arXiv preprint arXiv:2410.16662},
  year={2024}
}

@inproceedings{jiang2026large,
  title={When large multimodal models confront evolving knowledge: Challenges and explorations},
  author={Jiang, Kailin and Du, Yuntao and Ding, Yukai and Ren, Yuchen and Jiang, Ning and Gao, Zhi and Zheng, Zilong and Liu, Lei and Li, Bin and Li, Qing},
  booktitle={The Fourteenth International Conference on Learning Representations},
  year={2026}
}

@article{srinivasan2026benchmarking,
  title={Benchmarking large language models for ophthalmology (BELO): an expert-curated data set and evaluation framework for knowledge and reasoning},
  author={Srinivasan, Sahana and Ai, Xuguang and Lo, Thaddaeus Wai Soon and Gilson, Aidan and Zou, Minjie and Zou, Ke and Kim, Hyunjae and Yang, Mingjia and Pushpanathan, Krithi and Yew, Samantha Min Er and others},
  journal={Ophthalmology Science},
  volume={6},
  number={3},
  pages={101050},
  year={2026},
  publisher={Elsevier}
}

@article{zou2025benchmarking,
  title={Benchmarking next-generation reasoning-focused large language models in ophthalmology: a head-to-head evaluation on 5,888 items},
  author={Zou, Minjie and Srinivasan, Sahana and Lo, Thaddaeus Wai Soon and Zou, Ke and Yang, Gabriel Dawei and Ai, Xuguang and Kim, Hyunjae and Singer, Maxwell and Antaki, Fares and Li, Kelvin and others},
  journal={arXiv preprint arXiv:2504.11186},
  year={2025}
}

@article{chen2026visual,
  title={From visual question answering to intelligent AI agents in ophthalmology},
  author={Chen, Xiaolan and Chen, Ruoyu and Xu, Pusheng and Wan, Xiaojie and Zhang, Weiyi and Yan, Bingjie and Shang, Xianwen and He, Mingguang and Shi, Danli},
  journal={British Journal of Ophthalmology},
  volume={110},
  number={1},
  pages={1--7},
  year={2026},
  publisher={BMJ Publishing Group Ltd}
}

@article{zhou2026drvd,
  title={DrVD-Bench: Do Vision-Language Models Reason Like Human Doctors in Medical Image Diagnosis?},
  author={Zhou, Tianhong and Zhu, Yingtao and Xiao, Chuxi and Bian, Haiyang and Wei, Lei and Zhang, Xuegong and others},
  journal={Advances in Neural Information Processing Systems},
  volume={38},
  year={2026}
}

@article{arikan2025oct5k,
  title={OCT5k: A dataset of multi-disease and multi-graded annotations for retinal layers},
  author={Arikan, Mustafa and Willoughby, James and Ongun, Sevim and Sallo, Ferenc and Montesel, Andrea and Ahmed, Hend and Hagag, Ahmed and Book, Marius and Faatz, Henrik and Cicinelli, Maria Vittoria and others},
  journal={Scientific data},
  volume={12},
  number={1},
  pages={267},
  year={2025},
  publisher={Nature Publishing Group UK London}
}

@inproceedings{subramanian2022classification,
  title={Classification of retinal oct images using deep learning},
  author={Subramanian, Malliga and Shanmugavadivel, Kogilavani and Naren, Obuli Sai and Premkumar, K and Rankish, K},
  booktitle={2022 international conference on computer communication and informatics (ICCCI)},
  pages={1--7},
  year={2022},
  organization={IEEE}
}

@article{ye2023oimhs,
  title={Oimhs: An optical coherence tomography image dataset based on macular hole manual segmentation},
  author={Ye, Xin and He, Shucheng and Zhong, Xiaxing and Yu, Jiafeng and Yang, Shangchao and Shen, Yingjiao and Chen, Yiqi and Wang, Yaqi and Huang, Xingru and Shen, Lijun},
  journal={Scientific Data},
  volume={10},
  number={1},
  pages={769},
  year={2023},
  publisher={Nature Publishing Group UK London}
}

@article{hu2024amd,
  title={AMD-SD: an optical coherence tomography image dataset for wet AMD lesions segmentation},
  author={Hu, Yunwei and Gao, Yundi and Gao, Weihao and Luo, Wenbin and Yang, Zhongyi and Xiong, Fen and Chen, Zidan and Lin, Yucai and Xia, Xinjing and Yin, Xiaolong and others},
  journal={Scientific Data},
  volume={11},
  number={1},
  pages={1014},
  year={2024},
  publisher={Nature Publishing Group UK London}
}

@article{kulyabin2024octdl,
  title={Octdl: Optical coherence tomography dataset for image-based deep learning methods},
  author={Kulyabin, Mikhail and Zhdanov, Aleksei and Nikiforova, Anastasia and Stepichev, Andrey and Kuznetsova, Anna and Ronkin, Mikhail and Borisov, Vasilii and Bogachev, Alexander and Korotkich, Sergey and Constable, Paul A and others},
  journal={Scientific data},
  volume={11},
  number={1},
  pages={365},
  year={2024},
  publisher={Nature Publishing Group UK London}
}

@article{jbhi22-mmcamd,
  author={Weisen Wang and Xirong Li and Zhiyan Xu and Weihong Yu and Jianchun Zhao and Dayong Ding and Youxin Chen},
  journal={IEEE Journal of Biomedical and Health Informatics},
  title={Learning Two-Stream CNN for Multi-Modal Age-Related Macular Degeneration Categorization}, 
  year={2022},
  volume={26},
  number={8},
  pages={4111-4122},
  year={2022},
  doi={10.1109/JBHI.2022.3171523},
}

@inproceedings{fang2022dataset,
  title={Dataset and evaluation algorithm design for goals challenge},
  author={Fang, Huihui and Li, Fei and Fu, Huazhu and Wu, Junde and Zhang, Xiulan and Xu, Yanwu},
  booktitle={International Workshop on Ophthalmic Medical Image Analysis},
  pages={135--142},
  year={2022},
  organization={Springer}
}

@article{xun2023evidence,
  title={Evidence-based guidelines for diagnosis and treatment of diabetic retinopathy in China (2022)},
  author={Xun, X and Xiaoxin, L},
  journal={Chin. J. Ocular Fund. Dis},
  volume={2},
  pages={99--124},
  year={2023}
}

@article{vemulakonda2025age,
  title={Age-related macular degeneration preferred practice pattern{\textregistered}},
  author={Vemulakonda, G Atma and Bailey, Steven T and Kim, Stephen J and Kovach, Jaclyn L and Lim, Jennifer I and Ying, Gui-shuang and Flaxel, Christina J and others},
  journal={Ophthalmology},
  volume={132},
  number={4},
  pages={P1--P74},
  year={2025}
}

@article{lim2025diabetic,
  title={Diabetic retinopathy preferred practice pattern{\textregistered}},
  author={Lim, Jennifer I and Kim, Stephen J and Bailey, Steven T and Kovach, Jaclyn L and Vemulakonda, G Atma and Ying, Gui-shuang and Flaxel, Christina J},
  journal={Ophthalmology},
  volume={132},
  number={4},
  pages={P75--P162},
  year={2025},
  publisher={Elsevier}
}

@article{kim2025idiopathic,
  title={Idiopathic macular hole preferred practice Pattern{\textregistered}},
  author={Kim, Stephen J and Lim, Jennifer I and Bailey, Steven T and Kovach, Jaclyn L and Vemulakonda, G Atma and Ying, Gui-shuang and Flaxel, Christina J and others},
  journal={Ophthalmology},
  volume={132},
  number={4},
  pages={P234--P269},
  year={2025}
}

@article{kovach2025retinal,
  title={Retinal and ophthalmic artery occlusions preferred practice pattern{\textregistered}},
  author={Kovach, Jaclyn L and Bailey, Steven T and Kim, Stephen J and Lim, Jennifer I and Vemulakonda, G Atma and Ying, Gui-Shuang and Flaxel, Christina J and others},
  journal={Ophthalmology},
  volume={132},
  number={4},
  pages={P270--P302},
  year={2025}
}

@book{duker2021handbook,
title={Handbook of Retinal OCT: Optical Coherence Tomography E-Book},
author={Duker, Jay S and Waheed, Nadia K and Goldman, Darin},
year={2021},
publisher={Elsevier Health Sciences}
}

@inproceedings{liu2026amo,
  title={Amo-bench: Large language models still struggle in high school math competitions},
  author={Liu, Junlin and An, Shengnan and Zhou, Shuang and Ma, Dan and Lin, Yehao and Lv, Xinxuan and Wang, Xuanlin and Li, Xiaoyu and Wang, Ziwen and Cao, Xuezhi and others},
  booktitle={Findings of the Association for Computational Linguistics: ACL 2026},
  pages={2120--2137},
  year={2026}
}

@article{liu2026general365,
  title={General365: Benchmarking General Reasoning in Large Language Models Across Diverse and Challenging Tasks},
  author={Liu, Junlin and An, Shengnan and Zhou, Shuang and Ma, Dan and Luo, Shixiong and Xie, Ying and Zhang, Yuan and Yuan, Wenling and Zhou, Yifan and Li, Xiaoyu and others},
  journal={arXiv preprint arXiv:2604.11778},
  year={2026}
}

@article{jiang2025kore,
  title={KORE: Enhancing Knowledge Injection for Large Multimodal Models via Knowledge-Oriented Augmentations and Constraints},
  author={Jiang, Kailin and Jiang, Hongbo and Jiang, Ning and Gao, Zhi and Bi, Jinhe and Ren, Yuchen and Li, Bin and Du, Yuntao and Liu, Lei and Li, Qing},
  journal={arXiv preprint arXiv:2510.19316},
  year={2025}
}

@inproceedings{jiang2026mined,
  title={Mined: Probing and updating with multimodal time-sensitive knowledge for large multimodal models},
  author={Jiang, Kailin and Jiang, Ning and Du, Yuntao and Ren, Yuchen and Li, Yuchen and Gao, Yifan and Bi, Jinhe and Ma, Yunpu and Li, Bin and Liu, Lei and others},
  booktitle={Findings of the Association for Computational Linguistics: ACL 2026},
  pages={13766--13795},
  year={2026}
}

@article{peng2025can,
  title={Can Visual Input Be Compressed? A Visual Token Compression Benchmark for Large Multimodal Models},
  author={Peng, Tianfan and Du, Yuntao and Ji, Pengzhou and Dong, Shijie and Jiang, Kailin and Ma, Mingchuan and Tian, Yijun and Bi, Jinhe and Li, Qian and Du, Wei and others},
  journal={arXiv preprint arXiv:2511.02650},
  year={2025}
}

@inproceedings{jia2026benchmarking,
  title={Benchmarking multimodal knowledge conflict for large multimodal models},
  author={Jia, Yifan and Du, Yuntao and Jiang, Kailin and Liang, Yuyang and Ren, Qihan and Xin, Yi and Yang, Rui and Feng, Fenze and Chen, MingCai and Lu, Hengyang and others},
  booktitle={Proceedings of the AAAI Conference on Artificial Intelligence},
  volume={40},
  number={27},
  pages={22283--22291},
  year={2026}
}

@inproceedings{jiang2025mmke,
  title={Mmke-bench: A multimodal editing benchmark for diverse visual knowledge},
  author={Jiang, Kailin and Gao, Zhi and Shi, Chenrui and Zheng, Zilong and Qi, Siyuan and Li, Qing and others},
  booktitle={International Conference on Learning Representations},
  volume={2025},
  pages={526--555},
  year={2025}
}

@inproceedings{qi2025context,
  title={In-context editing: Learning knowledge from self-induced distributions},
  author={Qi, Siyuan and Yang, Bangcheng and Jiang, Kailin and Wang, Xiaobo and Li, Jiaqi and Zhong, Yifan and Yang, Yaodong and Zheng, Zilong},
  booktitle={International Conference on Learning Representations},
  volume={2025},
  pages={77563--77585},
  year={2025}
}


\end{document}